\documentclass[graybox]{svmult}

\usepackage[numbers]{natbib}

\usepackage{mathptmx}       
\usepackage{afterpage}       

\usepackage{helvet}         
\usepackage{courier}        
\usepackage{type1cm}        
\usepackage{makeidx}         
\usepackage{graphicx}        
\usepackage{multicol}        
\usepackage[bottom]{footmisc}
\usepackage{amsmath}
\usepackage{bm}
\usepackage{amssymb}
\usepackage{algorithm}
\usepackage{algpseudocode}
\usepackage{tikz}
\usepackage{listings}
\usepackage{multirow}

\makeindex             

\begin{document}

\title*{Machine learning and evolutionary techniques in interplanetary trajectory design}
\author{Dario Izzo \and Christopher Sprague \and Dharmesh Tailor}
\institute{Dario Izzo \at European Space Agency, Noordwijk, 2201 AZ, The Netherlands \email{dario.izzo@esa.int}
\and Christopher Iliffe Sprague \at KTH Royal Institute of Technology, Teknikringen 14, SE-100 44 Stockholm, Sweden \email{sprague@kth.se}
\and Dharmesh Vijay Tailor \at European Space Agency, Noordwijk, 2201 AZ, The Netherlands \email{dharmesh.tailor@live.co.uk}
}
%
%
\maketitle

\abstract*{
After providing a brief historical overview on the synergies between artificial intelligence research, in the areas of evolutionary computations and machine learning, and the optimal design of interplanetary trajectories, we propose and study the use of deep artificial neural networks to represent, on-board, the optimal guidance profile of an interplanetary mission. The results, limited to the chosen test case of an Earth-Mars orbital transfer, extend the findings made previously for landing scenarios and quadcopter dynamics, opening a new research area in interplanetary trajectory planning.
}

\abstract{
After providing a brief historical overview on the synergies between artificial intelligence research, in the areas of evolutionary computations and machine learning, and the optimal design of interplanetary trajectories, we propose and study the use of deep artificial neural networks to represent, on-board, the optimal guidance profile of an interplanetary mission. The results, limited to the chosen test case of an Earth-Mars orbital transfer, extend the findings made previously for landing scenarios and quadcopter dynamics, opening a new research area in interplanetary trajectory planning.
}

\section{Introduction}
\label{sec:1}
The use of Artificial Intelligence (AI) techniques for the design of interplanetary trajectories, in particular in their early design phases, has been proposed and studied by numerous scientists over the past few decades. Among the AI techniques deployed to help the optimisation of spacecraft trajectories are evolutionary algorithms~\cite{janin1985genetic,gage1995interplanetary, izzo2007search, lee2005multi, sentinella2009hybrid, vasile2010analysis, olds2007interplanetary, izzo2013search}, machine learning techniques~\cite{ampatzis2009machine, cassioli2012machine, mereta2017machine, izzo2016designing}, evolutionary neuro-controllers~\cite{dachwald2004low, bernd:2018}, tree search methods~\cite{grigoriev2013choosing, hennes2015interplanetary, simoes2017multi, izzo2016designing, izzo2013search} to only name a few widely studied methods. While the synergies between research in AI and in trajectory design were becoming increasingly apparent, the field of AI as a whole experienced a renaissance in the second decade of the millennium, delivering exciting developments that are of significance to many scientific fields. Certainly the area of machine learning is one that is powering such an AI renaissance, in particular with Deep Learning techniques \cite{schmidhuber2015deep} becoming fundamental to define new benchmarks in the most diverse applications. It thus seems likely that also the optimisation of interplanetary trajectories, already receptive of AI methods in general, is going to benefit from the vast amount of new knowledge being produced in the context of AI research. 

In this chapter we first briefly provide an overview on the state-of-the art of the use of evolutionary techniques (Section \S \ref{sec:2}) and machine learning techniques (Section \S \ref{sec:3}) when it comes to optimising interplanetary trajectories, and we then introduce a novel idea concerning the application of Deep Learning for the on-board representation of the optimal guidance profile for an interplanetary probe, extending previous work made on planetary landing scenarios~\cite{sanchez2016learning, sanchez2016real, schiavone2012autonomous}.

\section{Evolutionary algorithms for trajectory planning, an overview}
\label{sec:2}
Evolutionary algorithms are a class of global optimisation techniques that make use of heuristic rules, often inspired but not limited to natural paradigms such as Darwinian evolution, to search for optimal solutions in discontinuous, rugged landscapes. As such they are a good match to solve interplanetary trajectory optimisation problems where the planetary geometry defines a quite complex solution landscape already in simple cases. Already in 1985 a Genetic Algorithm (GA) \cite{janin1985genetic} was proposed and studied in the context of interplanetary trajectory optimisation concluding that 
\begin{quotation}
... a considerable effort is still needed for developing efficient schemes using genetic algorithms. However, they appear to offer an entirely original way for solving a large class of [interplanetary trajectory] global optimisation problems ...
\end{quotation}
Given the CPU power available at the time, it is only  natural that this pioneering work complained about efficiency. In the following decade the \lq\lq entirely original way\rq\rq\ was consolidated and more successful applications of GAs for both low-thrust propelled spacecraft \cite{rauwolf1996near} and impulsive thrust strategies \cite{gage1995interplanetary} were deployed. Studies on the use of some form of GA, also leveraging on the ever increasing computational power available, continued (see for example \cite{rogata2003guess, biesbroek2002optimization, gad2011hidden, deb2007interplanetary, elsayed2011ga}) showing how the original intuition, back in 1985, was one rich of consequences. It was indeed relatively consequential to substitute the GA with any other form of stochastic optimiser, so that in 2004 the European Space Agency performed a series of studies, in collaboration with academia, to test and compare several global optimisers on interplanetary trajectory problems \cite{ari1, ari2, izzo2007search}. It then became clear how other evolutionary algorithms were offering alternative, and in many ways superior, choices to help the preliminary phases of trajectory design. Among them, Differential Evolution (DE) and Particle Swarm Optimisation (PSO) were identified and benchmarked on several chemical propulsion problems opening the way to further independent confirmations of their performances \cite{vasile2010analysis, yao2017improved, olds2007interplanetary, sentinella2009hybrid, luo2005simulated, pontani2010particle}. Interestingly, in 2013, a self-adaptive Differential Evolution algorithm was used to plan the grand tour of the Jupiter moons that was awarded the golden Humies medal \cite{izzo2013search} for \lq\lq human-competitive results
produced by genetic and evolutionary computation\rq\rq. Many other evolutionary approaches have been proved to be of use by various researchers during the early 2000s, including Simulated Annealing (SA) \cite{luo2005simulated} and Ant Colony Optimisation (ACO) \cite{radice2006ant, ceriotti2010mga}. More recently, Covariance Matrix Evolutionary Strategy (CMA-ES) was shown to be potentially outperforming self-adaptive DE on a class of transfers \cite{izzo2014constraint}.  

While the no free lunch theorem \cite{wolpert1997no} guarantees that no evolutionary approach is better than any other on average, restricting the class of optimisation problems to those representing interplanetary transfers allows to identify the most useful approaches. To this end, a set of problems called GTOP database \cite{vinko2008global} (a sort of open trajectory gym) was created and made available to the scientific community and is still the subject of active research \cite{stracquadanio2011design, addis2011global, schlueter2014midaco, islam2012adaptive, cassioli2012machine, simoes2014self}. Some of the the interplanetary trajectory problems (i.e. Messenger and Cassini2) in the GTOP database were also used during the CEC2011 competition attracting the attention of the larger scientific community of evolutionary computations (see \cite{elsayed2011ga} for the competition winner).

Most of the research mentioned so far, considers continuous optimisation problems with a single objective and box constraints. In its most general case, though, the problem encountered in interplanetary trajectory design are multi-objective,  with nonlinear constraints and, possibly, integer decision variables. Genetic approaches to multi-objective trajectory design were benchmarked already in 2005 \cite{lee2005multi} followed by deeper studies on Non Dominated Sorting Genetic Algorithm (NSGA-II) that included also, as integer variables, the planetary fly-by sequence \cite{deb2007interplanetary}. Multi-objective versions of PSO have also been considered early on \cite{lavagna2007multi}. The more modern Multi Objective Evolutionary Algorithm by Decomposition (MOEA/D) was later identified as a most performing technique in applicable cases \cite{izzo2014constraint}, in the same work, several constraint handling techniques including co-evolution and immune systems were tested.

Whenever the the representation of an interplanetary trajectory requires integer and continuous variables, the resulting optimisation problem (a MINLP) typically becomes intractable also for evolutionary approaches and while some results have been obtained, for example using the ACO paradigm \cite{schlueter2013midaco, ceriotti2010mga}, a more convincing approach is to consider the continuous part and the integer part of the problem separately and architect some optimisation scheme tackling the two problems with different nested techniques (bi-level optimisation). Evolutionary approaches based on GAs \cite{izzo2015evolving, izzo2014gtoc5, englander2013automated} or ACO \cite{simoes2017multi} have been used in bi-level optimisation schemes, often coupled with smart tree search strategies such as Beam Search (BS) \cite{grigoriev2013choosing, simoes2017multi}, Monte Carlo Tree Search (MCTS) \cite{hennes2015interplanetary}, Multi Objective Beam Search (MOBS) \cite{izzo2016designing} or Lazy Race Tree Search (LRTS) \cite{izzo2013search} to take care of the integer part.

Evolutionary approaches to interplanetary trajectory planning have proven their worth beyond any criticism and while they are still not as widely used by the aerospace industry as they could, it is likely that we will see a larger penetration in the industrial sector in the upcoming years, as the interest in Artificial Intelligence methods powered by Deep Learning and Machine Learning will put also evolutionary techniques in the spotlight.

\section{Machine Learning and interplanetary trajectories}
\label{sec:3}
The use of machine learning (ML) algorithms to aid the design of interplanetary trajectory is not as widely researched as that of evolutionary techniques and is limited to fewer works. The reasons are to be found in the less obvious applicability of these methods to the problems encountered in the design of interplanetary trajectories and in the lack of data sets produced and made available by the aerospace community. In this section we try to summarise the ideas that have so far been proposed.

During the optimisation of an interplanetary transfer, as in any optimisation task, a large number of solutions are computed and assessed to inform the search for better candidates. Typically all these design points are discarded and lost after they have been used to produce new promising search directions. The idea of applying supervised learning on such a data set in order to build a model improving the further selection of initial guesses to guide successive evolutionary runs was proposed and tested \cite{cassioli2012machine} on some of the trajectory problems in the GTOP database \cite{vinko2008global} using Support Vector Machines (SVM). Following a similar reasoning, a ML model can be trained, using the points sampled during the evolution, in order to construct a surrogate model of the trajectory worth, which then avoids expensive evaluations of the objective function \cite{ampatzis2009machine}. Building surrogate models is particularly relevant when the interplanetary mission fitness requires a high degree of computational resources such as in the case of optimal low-thrust transfers. In that case a surrogate model approximating the final optimal transfer mass enables to quickly search for good launch and arrival epochs, as well as planetary body sequences, e.g. in the case of multiple asteroid or debris rendezvous missions \cite{hennes2016fast, mereta2017machine}. Unsupervised learning techniques such as clustering or nearest neighbours have also been used to select the target of transfers in multiple asteroid rendezvous missions, upon proper definition of a metric coping with the orbital non linearities \cite{izzo2016designing}, or to define new box bounds and hence focus successive evolutionary runs in promising areas of the search space (cluster pruning \cite{izzo2010global}).

A second field where ML algorithms have been used in the context of research in interplanetary trajectory design methods, is that of the on-board representation of the optimal guidance profile. Already in 2004 \cite{dachwald2004low} the idea was put forward of using machine learning to learn a representation of the optimal spacecraft guidance profile. In later years the technique was studied exclusively in the context of evolutionary neuro-controllers and optimal control. More recently, neuro evolution has been substituted with supervised learning (back-propagation), the artificial neural network structure has become deeper, and focus has been shifted to the on-board real time computation of guidance profiles \cite{sanchez2016real, sanchez2016learning, schiavone2012autonomous}. These last works have all been limited, to landing problems and the applicability of these ideas to interplanetary trajectories has not been studied. In the next section we will try to contribute closing this gap showing how the optimal guidance profile of a phase-less Earth-Mars transfer can also be satisfactorily represented by a deep artificial neural network.

\section{On-board optimal guidance via a deep network}\label{sec:4}
In planetary landing problems, the possibility to generate optimal guidance profiles on-board has been studied and suggested to significantly enhance a vehicle’s landing accuracy \cite{de2012real, dueri2016customized}. The resulting algorithms need to run on a radiation-hardened flight processor, that is on a significantly slower processor than those available on modern desktops and one having significant architectural differences. As a rule of thumb, a space qualified processor is an order of magnitude slower than the average desktop processor. The resulting CPU load on the spacecraft has been estimated in a practical scenario \cite{dueri2016customized} to be 0.7s for the computation of one optimal action. To the same end, an alternative is offered by deep feed-forward neural networks trained on the ground to approximate the optimal control and used on-board in real time. The on-board computational effort associated to this architecture is that of one forward pass of the network which, though deep, is not necessarily large \cite{sanchez2016learning, sanchez2016real}. Extending on previous studies on landing and quadcopter dynamics, the feasibility of such a scheme for interplanetary low-thrust transfers is studied here. The same overall scheme is used \cite{sanchez2016learning}:
\begin{svgraybox}
\begin{itemize}
    \item Step 1: We solve thousands of optimal control problems using Pontryagin's maximum principle. We store, with some time sampling, the obtained solutions in a data set.
    \item Step 2: We train deep feed-forward neural networks on the data set to learn the optimal control structure.
    \item Step 3: We use, on-board, the trained network to compute the optimal feedback.
\end{itemize}
\end{svgraybox}
With respect to the previously studied case of planetary landings the methodology is, essentially, unchanged and it is of great interest to see how it can be applied on a radically different problem having larger dimensionality, different non linearities and, arguably, a more complex structure.

\subsubsection*{A short note on notation}
We use boldface symbols (e.g. $\boldsymbol{\lambda}, \mathbf r$, etc.) to denote vector quantities, while scalars or the vector norm will be indicated by normal symbols (e.g. $\lambda, r$, etc.). Unit vectors will be indicated with a small hat (e.g. $\mathbf{\hat i}$) and time derivatives with a dot (e.g. $\dot x$). We will sometime use the asterisk as a superscript to indicate optimal quantities (e.g. $u^*$).

\subsection {Spacecraft Dynamics}
The motion of a spacecraft equipped with a constant specific impulse nuclear electric low-thrust propulsion system can be described in some heliocentric, inertial, frame using Cartesian coordinates by the equations:
\begin{equation}\label{eq:eom}
\begin{array}{l}
    \dot {\mathbf r} = \mathbf v\\
    \dot {\mathbf v} = -\frac{\mu}{r^3}\mathbf r + c_1 \frac {u(t)}m \mathbf{\hat i_u} (t) \\
    \dot m = - c_2 u(t)
\end{array}
\end{equation}
where $\mathbf r$, $\mathbf v$ and $m$ denote the spacecraft position, velocity and mass (also denoted as spacecraft state $\mathbf x$), $\mu$ is the gravitational parameter of the Sun, $c_1$ the maximum thrust achievable by the onboard propulsion system and $c_2 = c_1 / I_{sp} g_0$. We have denoted with $I_{sp}$ the propulsion specific impulse and with $g_0$ the Earth gravity constant at sea level. The control variables $u(t)$ and $\mathbf{\hat i_u}(t)$ (or $\mathbf u(t)$ for brevity) describe the thrust level (here also called throttle) and its direction and are constrained as follows $|u(t)| \le 1$ and $|\mathbf{\hat i_u}(t)| = 1, \forall t \in [t_1, t_2]$.

We are interested in the problem of finding the controls $u(t)$ and $\mathbf{\hat i_u}(t)$ to minimise the functional:
\begin{equation}\label{eq:J}
J(t_1, t_2, u(t)) = \alpha \int_{t_1}^{t_2} u(t) dt + (1 - \alpha) \int_{t_1}^{t_2} u^2(t) dt
\end{equation}
and able to steer the spacecraft from some initial point $x_1 \in \mathcal S_1$ to some final point $x_2 \in \mathcal S_2$ where the sets $\mathcal S_1$ and $\mathcal S_2$ are some closed subsets (hypersurfaces) of the state space. Note that the functional $J$ is parameterised by the continuation parameter $\alpha \ \in [0,1]$ which weights two contributions corresponding to mass optimality ($\alpha = 1$) and quadratic control optimality ($\alpha = 0 $). 

We now derive the, known, necessary optimality conditions for the above stated problem applying the maximum principle from Pontryagin (note that we have stated a minimisation problem, hence the conditions are actually slightly different from the ones originally derived in Pontryagin work \cite{pontryagin}).

\subsection {Minimisation of the Hamiltonian $\mathcal H$}
We start by introducing seven auxiliary functions defined in $t \in [t_1, t_2]$, the co-states, which we will denote with $\boldsymbol \lambda_r$,$\boldsymbol \lambda_v$ and $ \lambda_m$, or, for brevity, $\boldsymbol \lambda$. We then introduce the Hamiltonian function:
\begin{equation}
\label{eq:hamiltonian}
\mathcal H(\mathbf x, \boldsymbol \lambda, \mathbf u) = \boldsymbol \lambda_r \cdot \mathbf v + \boldsymbol \lambda_v \cdot \left(-\frac{\mu}{r^3}\mathbf r + c_1 \frac {u}m \mathbf{\hat i_u}\right) - \lambda_m c_2 u 
 + \alpha u + (1-\alpha) u^2
\end{equation}
which, following Pontryagin theory, needs to be minimised by our controls along an optimal trajectory. Isolating the relevant part of the Hamiltonian that depends on the control $\mathbf{\hat i_u}$ we have that $\mathcal H = ... + c_1 \frac um \boldsymbol \lambda_v \cdot \mathbf{\hat i_u} + ... $, which allows to conclude that, since $m, u$ and $c_1$ are all positive numbers, the thrust direction must be in the opposite direction of $\boldsymbol \lambda_v$ for $\mathcal H$ to be minimised, more formally:
\begin{equation}\label{eq:primer}
\mathbf{\hat i_u}^* = -\frac{\boldsymbol \lambda_v}{\lambda_v}
\end{equation}
Substituting this expression back into the Hamiltonian we get:
\begin{equation*}
\label{eq:hamiltonian}
\mathcal H(\mathbf x, \boldsymbol \lambda, \mathbf u) = \boldsymbol \lambda_r \cdot \mathbf v 
- \frac{\mu}{r^3} \boldsymbol \lambda_v \cdot \mathbf r  - c_1 \frac um \lambda_v
- \lambda_m c_2 u  + \alpha u + (1-\alpha) u^2
\end{equation*}
which we now need to minimise with respect to $u$. Isolating the relevant part of $\mathcal H$, i.e. the terms that depend on $u$, we have:
$$
\mathcal H = ... + (1-\alpha) u^2 + u \left( \alpha  - \frac {c_1}m \lambda_v - \lambda_m c_2 \right) + ...
$$
which is a convex parabola that will take its minimal value in $u = \frac{\frac {c_1}m \lambda_v + \lambda_m c_2 - \alpha }{ 2 (1-\alpha)}$. Since $u \in [0,1]$, we get the final expression for an optimal $u^*$:
\begin{equation}\label{eq:optimalu}
u^* = \min\left[\max\left( \frac{\frac {c_1}m \lambda_v + \lambda_m c_2 - \alpha }{ 2 (1-\alpha)}, 0 \right)  ,1\right]
\end{equation}
Note that in the corner case $\alpha = 1$, which correspond to a mass optimal control, the above expression results to be singular and the minimiser of the Hamiltonian can be conveniently rewritten introducing the switching function,
$$
S_{sw} =  c_1\lambda_v + m c_2 \lambda_m - m \alpha
$$
as:
\begin{equation}\label{eq:massoptimalu}
u^* = \left\{\begin{array}{ll}
1 & \mbox{if} \quad S_{sw} > 0 \\
0 & \mbox{if} \quad S_{sw} < 0 \\
\end{array}\right.
\end{equation}

\subsection {The costate equations}
The states $\mathbf x(t)$ and co-states $\boldsymbol \lambda(t)$ must be a solution to the set of differential equations that are elegantly written using the Hamiltonian formalism as:
$$
\begin{array}{ll}
\dot{\mathbf r} = \frac{\partial H}{\partial \boldsymbol \lambda_r}, & \dot{\boldsymbol \lambda}_r = - \frac{\partial H}{\partial \mathbf r} \\
\dot{\mathbf v} = \frac{\partial H}{\partial \boldsymbol \lambda_v}, & \dot{\boldsymbol \lambda}_v = - \frac{\partial H}{\partial \mathbf v}  \\
\dot{m} = \frac{\partial H}{\partial \lambda_m}, & \dot{\lambda}_m = - \frac{\partial H}{\partial m}  \\
\end{array}
$$
while for the first three equations it is trivial to get back the expressions in Eq.(\ref{eq:eom}), the co-states equations do require some extra work, in particular when deriving the gravity gradient. 
Let us then start deriving the co-state equations by computing $\frac{\partial H}{\partial \mathbf r}$ starting from the expression in Eq.(\ref{eq:hamiltonian}). We have
$$
\frac{\partial H}{\partial \mathbf r} = - \boldsymbol \lambda_v \cdot \nabla \left(\frac{\mu}{r^3}\mathbf r\right)
$$
where the symbol $\nabla$ denotes the gradient operator. Its easier to compute the expression regrouping as follows:
\begin{eqnarray*}
-\frac 1\mu\frac{\partial H}{\partial \mathbf r}& = & \nabla \left( \boldsymbol \lambda_v \cdot \frac{1}{r^3}\mathbf r\right) = \\
& = & \frac{1}{r^3} \nabla\left( \boldsymbol \lambda_v \cdot \mathbf r \right) + \left(\boldsymbol \lambda_v  \cdot \mathbf r\right) \nabla \left(\frac 1{r^3}\right) = \\
& = & \frac{\boldsymbol \lambda_v}{r^3} - 3 \left(\boldsymbol \lambda_v  \cdot \mathbf r\right) \frac{ \nabla r }{r^4}= \\
& = & \frac{\boldsymbol \lambda_v}{r^3} - 3 \left(\boldsymbol \lambda_v  \cdot \mathbf r\right) \frac{\mathbf r}{r^5}
\end{eqnarray*}
We may thus write the corresponding Hamilton equation as: 
$$
\dot {\boldsymbol \lambda}_r = -\frac{\partial H}{\partial \mathbf r} = \mu \frac{\boldsymbol \lambda_v}{r^3} - 3 \mu\left(\boldsymbol \lambda_v  \cdot \mathbf r\right) \frac{\mathbf r}{r^5} 
$$
The remaining two equations are then also easily obtained as:
$$
\dot {\boldsymbol \lambda}_v = -\frac{\partial H}{\partial \mathbf v} = - \boldsymbol \lambda_r 
$$
and:
$$
\dot { \lambda}_m = - \frac{\partial H}{\partial m} =  c_1 u \frac{\boldsymbol \lambda_v \cdot \mathbf{\hat i_u}}{m^2}
$$
The final system of differential equations describing the state and co-states evolution along an optimal trajectory may be now summarised:
\begin{equation}\label{eq:eom_pontryagin}
\left\{
\begin{array}{l}
    \dot {\mathbf r} = \mathbf v\\
    \dot {\mathbf v} = -\frac{\mu}{r^3}\mathbf r - c_1 \frac {u^*(t)}m \frac{\boldsymbol \lambda_v}{\lambda_v} \\
    \dot m = - c_2 u^*(t) \\
    \dot {\boldsymbol \lambda}_r = \mu \frac{\boldsymbol \lambda_v}{r^3} - 3 \mu\left(\boldsymbol \lambda_v  \cdot \mathbf r\right) \frac{\mathbf r}{r^5} \\
     \dot {\boldsymbol \lambda}_v = - \boldsymbol \lambda_r  \\
    \dot { \lambda}_m = - c_1 u^*(t) \frac{ \lambda_v }{m^2}
\end{array}
\right.
\end{equation}

\subsection {The two-point boundary value problem}
Following the maximum principle, we know that there exist initial values for the co-states $\boldsymbol \lambda(t_1) = \boldsymbol \lambda_1$ such that an optimal trajectory steering the system from a starting state $\mathbf x_1 \in \mathcal S_1$ to a final state $\mathbf x_2 \in \mathcal S_2$ will be found integrating the system of equations stated in Eq.(\ref{eq:eom_pontryagin}) from the initial condition $\mathbf x_1, \boldsymbol \lambda_1$, for a time $t_2 - t_1$ leading to a final condition $\mathbf x_2, \boldsymbol \lambda_2$. Since we also want to find optimal values for $\mathbf x_1 \in \mathcal S_1$,  $\mathbf x_2 \in \mathcal S_2$ and $t_2$ we need some additional conditions. The problem of finding the optimal $t_2$ ($t_1$ can be assumed to be 0 as our system is autonomous) is solved by adding a condition on the Hamiltonian:
\begin{equation}\label{eq:transv_time}
\mathcal H(\mathbf x(t_2), \boldsymbol \lambda(t_2), \mathbf u^*(t_2)) = 0
\end{equation}
To optimally select $\mathbf x_1 \in \mathcal S_1$ and $\mathbf x_2 \in \mathcal S_2$, instead, we will need to add some conditions, called transversality conditions and derived in the next section.

\subsection {Transversality conditions}
Transversality conditions ensure that the initial and final states are selected optimally in the allowed sets $\mathcal S_1$ and $\mathcal S_2$. They can be elegantly written introducing the vectors $\boldsymbol \theta_1$ and $\boldsymbol \theta_2$ belonging to the hyperplane tangent to the hypersurfaces $\mathcal S_1$ and $\mathcal S_2$ in $\mathbf x_1$ and $\mathbf x_2$. For all such vectors it must be that:
$$
\boldsymbol \lambda(t_1) \cdot \boldsymbol \theta_1 = 0
$$
and equivalently,
$$
\boldsymbol \lambda(t_2) \cdot \boldsymbol \theta_2 = 0
$$
Let us now derive explicitly these relations in the case of the interplanetary transfer dynamics. We start focusing on the terminal condition $\mathbf x_2 \in \mathcal S_2$. Typically the final mass of the spacecraft is not specified and we may, in this case, express $\mathcal S_2$ as $\mathbf x_2 = [\mathbf r_2, \mathbf v_2, c]$ with $c$ being a free parameter and $\mathbf r_2, \mathbf v_2$, being at this stage considered as fixed. Trivially, in this case, $\boldsymbol \theta_2 = [\mathbf 0,\mathbf 0,\theta_2]$ and the transversality condition for a free final mass is thus:
\begin{equation}\label{eq:transversality_mass}
\boldsymbol \lambda_m(t_2) = 0
\end{equation}
More work is needed to derive transversality conditions in the case where also $\mathbf r_2, \mathbf v_2$ are not fixed but constrained to belong to some manifold. A typical situation would be, for example, to leave some of the final Keplerian orbital elements as free. In this case, in order to find 
$\boldsymbol \theta_2$, i.e. the vector tangent to the defined manifold, we need to write a parameterisation of such manifold, that is an expression of  $\mathbf r_2$ and  $\mathbf v_2$ as a function of the desired orbital parameter. Using the Keplerian elements $a,e,i,\omega,\Omega, E$ we may use well know relations to establish such a parameterisation:
\begin{equation}\label{eq:parametrization}
\begin{array}{l}
\mathbf r_2 = \mathbf R \left[\begin{array}{c}
    a(\cos E - e)   \\
    a\sqrt{1-e^2}\sin E \\
    0
\end{array}\right]\\
\mathbf v_2 = \sqrt{\frac{\mu}{a^3}}\frac{1}{1-e\cos E}\mathbf R \left[\begin{array}{c}
    - a\sin E   \\
    a\sqrt{1-e^2}\cos E \\
    0
\end{array}\right]
\end{array}
\end{equation}
where $\mathbf R$ is the rotation matrix from the orbital reference frame to the inertial defined as:
$$
\mathbf R = \left[
\begin{array}{ccc}
  \cos\Omega\cos\omega - \sin\Omega\sin\omega\cos i   & -\cos\Omega\sin\omega - \sin\Omega\cos\omega\cos i & \sin\Omega\sin i\\
   \sin\Omega\cos\omega+\cos\omega\sin\omega\cos i  &- \sin\Omega\sin\omega+\cos\Omega\cos\omega\cos i & -\cos\Omega\sin i\\
   \sin\omega\sin i  & \cos\omega\cos i& \cos i
\end{array}
\right]
$$
the tangent vector to the resulting manifold will then be simply obtained deriving the above expression with respect to the chosen parameter.

For example, let us take the case of a free final anomaly. This corresponds to a transfer to some final target orbit regardless of the phase. Deriving the above expressions with respect to $E$, or in this case equivalently $t$, we get the transversality condition:
\begin{equation}\label{eq:transversality_anomaly}
\boldsymbol\lambda_r \cdot \mathbf v_2 - \boldsymbol\lambda_v \cdot \frac{\mu}{r_2^3}\mathbf r_2 = 0
\end{equation}
Similarly if we leave the semi-major axis $a$ as free, the conditions is easily obtained as:
\begin{equation}\label{eq:transversality_anomaly}
2\boldsymbol\lambda_r \cdot \mathbf r_2 - \boldsymbol\lambda_v \cdot  \mathbf v_2 = 0
\end{equation}
Similar expressions are derived in \cite{pan2013reduced} using true anomaly and not the eccentric anomaly and using a different method.

\section{Test case description (nominal trajectories)}
\label{sec:5.1}
As a test case, we consider a low-thrust Earth-Mars orbital transfer. A spacecraft having mass $m_0 = 1000$ (kg) and equipped with a propulsion system able to deliver a constant $T_{max} = 0.3$ (N) with a specific impulse of $I_{sp} = 2500$ (s) starts its transfer from Earth's orbit to reach, after a time of flight $\Delta t$, Mars' orbit. Both starting and target orbits are considered Keplerian. The optimal transfer strategy is found by solving the resulting TPBVP where we seek the time of flight $\Delta t$, departure eccentric anomaly $E_0$, arrival eccentric anomaly $E_f$ and departure co-state variables $\lambda_0$ that satisfy the boundary conditions, the dynamics Eq.(\ref{eq:eom_pontryagin}), the transversality condition on free time Eq.(\ref{eq:transv_time}), free mass Eq.(\ref{eq:transversality_mass}) and free anomalies Eq.(\ref{eq:transversality_anomaly}).

\begin{figure}[t]
\sidecaption
\includegraphics[width=0.8\linewidth]{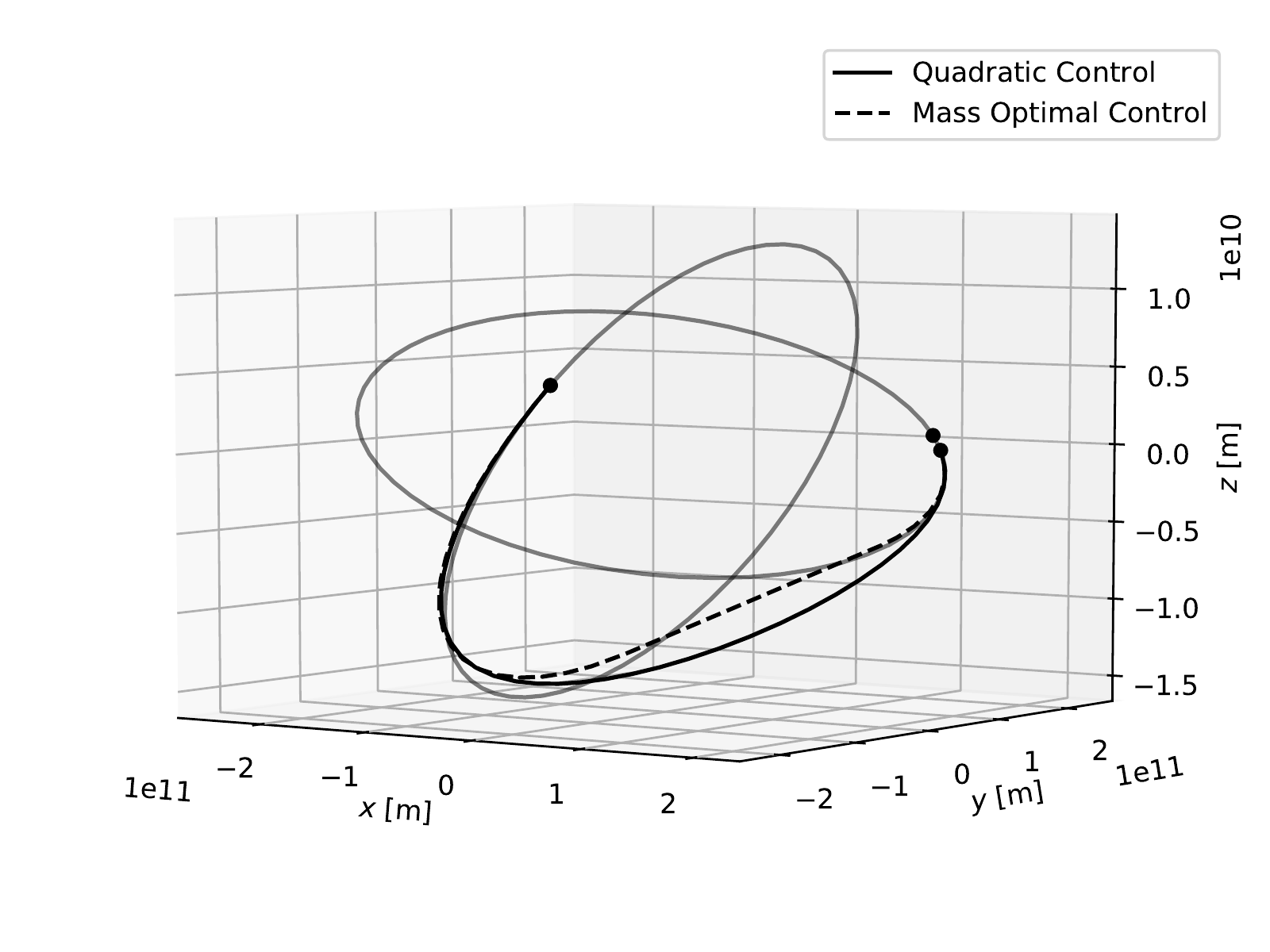}
\caption{The two nominal trajectories considered.}
\label{fig:nominal}
\end{figure}

The TPBVP is solved first for quadratic control optimality (QOC - $\alpha=0$). Once the optimal trajectory is found, the problem is then solved with gradually increasing $\alpha$ until mass-optimal control (MOC) is achieved $\alpha = 1$, at which point both solutions (nominal trajectories visualised in Figure \ref{fig:nominal}) are stored as:
$$
\mathbf{T}_{nom} = \left\{(\mathbf{x}, \boldsymbol{\lambda}, t)_{nom,j}\right\}, j = 0, \dots, J_{nom}
$$
where the grid points are determined by the adaptive integrator used. This will result in more points allocated in areas where the dynamics gradient is higher, which are also areas where we would want to have more training samples and is thus considered as an appropriate mechanism.

\section {Generating the QOC data set}
As we want to train a deep neural architecture to represent the optimal control around our nominal trajectories, we need to generate a large data set describing the functional relationship that is to be learned. Such a functional relationship is the optimal state feedback ($\mathbf x \to \mathbf u^*$) which can be computed along single trajectories neighbouring the nominal ones at the cost of solving the resulting TPBVPs.
In order to make use of previously computed solutions (starting from the nominal trajectories) to help the convergence of the TPBVP solver, a continuation technique is then employed where new TPBVPs are created using as initial states those of the nominal trajectories perturbed along a random walk and as initial co-states those computed from previous steps. This allows to generate efficiently the database continuing one starting solution, i.e. the nominal trajectory. The TPBVP is solved by means of a simple single shooting method. The pseudo algorithm used to fill the data-set is given in Algorithm \ref{alg:randomwalk} which is run starting from 10 different points equally spaced along the nominal trajectory and using $\alpha = 0$. Let the final data-set
\begin{equation}\label{eq:database}
\mathcal{T}_{\mbox{QOC}} = \left\{\mathbf{T}_i\right\},  i = 0,\dots,I
\end{equation}
contain $I$ trajectories
\begin{equation}\label{eq:trajectory}
\mathbf{T}_{i} = \left\{(\mathbf{x}, \boldsymbol{\lambda}, t)_{i,j}\right\},  j = 0, \dots, J_i
\end{equation}
sampled in $J_i$ nodes (as determined by the outcome of an adaptive step numerical integration) 
each recording the states $\mathbf{x}$, the co-states $\boldsymbol{\lambda}$, and the times $t$. We refer to this data set as QOC (quadratic optimal control).

\begin{figure}[t]
\sidecaption
\includegraphics[width=0.8\linewidth]{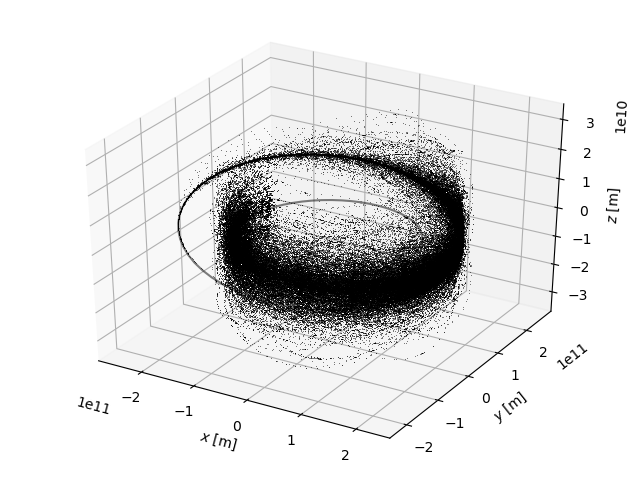}
\caption{Visualisation of the MOC data set containing $308,579$ optimal state-control pairs around the nominal trajectory.}
\label{fig:wholedata}
\end{figure}

\afterpage{%
\begin{algorithm}[t]
\caption{Random Walk}
\label{alg:randomwalk}
\begin{algorithmic}[1]
\Procedure {random walk}{$\mathbf{x}_{nom,j}, \boldsymbol{\lambda}_{nom,j}$, $\Delta t_{nom,j}$, $E_{f_{nom}}$, $\alpha$, $\overline \gamma$, $n$}
\State $\mathcal{T} \gets \{\}$ 
\Comment{Set of optimal trajectories}
\State $\mathbf{x}_0 \gets \mathbf{x}_{nom,j}$
\Comment{Set initial state as nominal}
\State $\boldsymbol{\lambda}_0 \gets \boldsymbol{\lambda}_{nom,j}$
\Comment{Set initial co-state as nominal}
\State $\Delta t \gets \Delta t_{nom,j}$
\Comment{Set time of flight as nominal} 
\State $E_{f} \gets E_{f_{nom}}$
\Comment{Set arrival eccentric anomaly as nominal} 
\State $\gamma \gets \overline \gamma$ 
\Comment{Nominal perturbation percentage}
\For{$i \gets 1, \dots, n$} \Comment{$n$ random perturbations}
    \State $\mathbf{\beta} \gets U(-1,1) \in\mathbb{R}^7$ 
    \Comment{Vector of random uniformly distributed numbers}
    \State $\mathbf{x}_1 \gets \mathbf{x}_0 + \mathbf{x}_0 \odot \mathbf{\beta}\gamma$ 
    \Comment{Perturb state in random direction by $\gamma$}
    \State $(\boldsymbol{\lambda}_1, \Delta t_1, E_{f_1}) \gets \text{TPBVP}(\mathbf{x}_1, \boldsymbol{\lambda}_0, \Delta t, E_{f}, \alpha)$
    \If{$\text{successful}$} \Comment{If the TPBVP is successfully solved}
        \State Update $\mathcal{T}$
        \Comment{Save successful trajectory}
        \State $\mathbf{x}_0 \gets \mathbf{x}_1$
        \Comment{Accept perturbed state}
        \State $\boldsymbol{\lambda}_0 \gets \boldsymbol{\lambda}_1$
        \Comment{Accept new computed co-states}
        \State $\Delta t_0 \gets \Delta t_1$
        \Comment{Accept new computed time of flight}
        \State $E_{f_0} \gets E_{f_1}$
        \Comment{Accept new computed eccentric anomaly}
        \State $\gamma \gets (\gamma + \overline \gamma)/2$
        \Comment{Increase perturbation size}
    \Else \Comment{If solution is unfeasible}
        \State $\gamma \gets \gamma/2$
        \Comment{Decrease perturbation size and solve again}
    \EndIf
\EndFor
\State\Return{$\mathcal{T}$}
\Comment{Return set of optimal trajectories}
\EndProcedure
\end{algorithmic}
\end{algorithm}

\begin{algorithm}[h]
\caption{Homotopy}
\label{alg:homo}
\begin{algorithmic}[1]
\Procedure {homotopy}{$\mathbf{x}_{QOC, 0}, \boldsymbol{\lambda}_{QOC, 0}, \Delta t_{QOC}, E_{f_{QOC}}, \alpha_{\text{tol}}$}
\State $\alpha \gets 1$
\Comment{Try to solve for mass-optimal at first}
\Loop
    \State $(\boldsymbol{\lambda}, \Delta t, E_f) \gets \text{TPBVP}(\mathbf{x}_{QOC, 0}, \boldsymbol{\lambda}_{QOC, 0}, \Delta t_{QOC}, E_{f_{QOC}}, \alpha)$
    \If{$\text{successful}$} \Comment{If the TPBVP is successfully solved}
        \State $\alpha^\star \gets \alpha$
        \Comment{Current best $\alpha$}
        \State $(\boldsymbol{\lambda}_{QOC, 0}, \Delta t_{QOC}, E_{f_{QOC}})
        \gets (\boldsymbol{\lambda}, \Delta t, E_f)$
        \Comment{$\mathbf{x}_0^\star$ doesn't change}
        \If{$\alpha < \alpha_{\text{tol}}$}
            \State $\alpha \gets (1+\alpha)/2$ 
            \Comment{Increase $\alpha$}
        \ElsIf{$\alpha \geq \alpha_{\text{tol}}$} \Comment{If $\alpha$ is close to 1}
            \State $\alpha \gets 1$
        \ElsIf{$\alpha = 1$} \Comment{If trajectory is feasible and mass-optimal}
            \State \Return{$\mathbf{T}$}
            \Comment{Break the loop and return trajectory}
        \EndIf
        
    \Else
        \State $\alpha \gets (\alpha + \alpha^\star)/2$
        \Comment{Decrease $\alpha$ and retry}
    \EndIf
    
\EndLoop
\EndProcedure
\end{algorithmic}
\end{algorithm}
\clearpage
}

\subsection{Generating the MOC data set}
A second data set containing mass optimal trajectories is obtained via continuation over the parameter $\alpha$ (homotopy). Iterating through the initial states and co-states in $\mathcal{T}_{\mbox{QOC}}$, an attempt is made to solve, starting from that guess, directly a mass-optimal control ($\alpha=1$). If the resulting trajectory is feasible, it is stored and the next data set entry is considered. If the trajectory is not feasible, the homotopy parameter is decreased, and the optimisation is reattempted. If successful, the homotopy parameter is saved as the current best and its current value is increased. If, instead, the TPBVP solver is not converging, the homotopy parameter is decreased, and the algorithm continues. The pseudo code describing the homotopy method is outlined in Algorithm \ref{alg:homo} and results in a new data set $\mathcal{T}_{\mbox{MOC}}$ containing mass optimal trajectories and visualised in Figure \ref{fig:wholedata}.

\section{Learning details and network architecture}
We consider for both of the data-sets, $\mathcal{T}_{\mbox{MOC}}$ and $\mathcal{T}_{\mbox{QOC}}$, the optimal state-action pairs ($\mathbf x, \mathbf u$) and we build a deep model of the relation $\mathbf u(\mathbf x)$, i.e. the optimal state feedback. The dimension of the values we have to model is $D=3$ as they represent a three dimensional vector (i.e. the thrust vector). We choose to represent such a vector in its polar coordinates so that:
$$
\mathbf u = u \left[
\sin{\theta}\cos{\phi},
\sin{\theta}\sin{\pi},
\cos{\theta}
\right]
$$
where $\theta$ is the polar angle, $\phi$ is the azimuth angle and $u$ is the throttle magnitude. 

In previous work \cite{sanchez2016real}, it was found that a feedforward, fully-connected neural network with 3 hidden layers and 32 units/layer could satisfactorily represent the optimal guidance profile for a problem with simpler dynamics.
Starting from that knowledge, and from the fact that we are here considering a higher dimensional case with a higher degree of complexity, we experimented with deeper and wider networks.
The result of our investigation indicated that peak performance could be obtained by a neural network with four hidden layers and 200 units per layer.
Larger networks either matched this performance or performed worse.
We settled on the Rectified Linear Unit (ReLU) as the activation function for the hidden layers.
In addition, we used a hyperbolic tangent activation function for the output layer.
Thus we scaled the targets in the dataset to the range $[-1,1]$.
We initialised the network's weights using the heuristic in \cite{glorot2010understanding} in which random values are sampled from a uniform distribution close to zero.
Biases were initialised to zero.
We also normalised our input data such that features had a mean of zero and standard deviation of one; this helped to speed up the optimisation process.
We used the Adam training algorithm \cite{DBLP:journals/corr/KingmaB14}, a variant of stochastic gradient descent.
Weights were updated based on a minibatch of 64 training examples.
For the performance metric we use the commonly used mean squared error loss function (MSE). 
A 10\% split is used to define training and validation data.
We used an adaptive learning rate starting from an initial value of $10^{-3}$.
The learning rate was reduced by a factor of 10 when we observed a plateau in the training loss for a period greater than 10 epochs.
Furthermore, we stopped training when we observed a plateau in the training loss for greater than 50 epochs (`early stopping').
The plateau was defined to be a decrease in loss less than $10^{-4}$.
We experimented with different values for the hyperparameters described above before settling on the final configuration.

\begin{table}[t]
\centering
\begin{tabular}[t]{|l||l|l|l||l|}
\hline
\multirow{2}{*}{Network} & \multicolumn{3}{c||}{MSE} & \multirow{2}{*}{Epochs} \\ \cline{2-4}
                         & \multicolumn{1}{c|}{u} & \multicolumn{1}{c|}{$\phi$} & \multicolumn{1}{c||}{$\theta$} & \\ \hline
$\mathcal N^{QOC}_{u, \phi, \theta}$ &0.0178 &\bf{0.0690} &\bf{0.0870} &453 \\ \hline
$\mathcal N^{QOC}_{\phi, \theta}$ &\_ &0.0755 &0.0907 &461 \\ \hline
$\mathcal N^{QOC}_{u}$ &\bf{0.0075} &\_ &\_ &247\\ \hline
$\mathcal N^{QOC}_{\phi}$ &\_ &0.0698 &\_ &394 \\ \hline
$\mathcal N^{QOC}_{\theta}$ &\_ &\_ &0.0976 &503 \\ \hline
\end{tabular}
\quad
\begin{tabular}[t]{|l||l|l|l||l|}
\hline
\multirow{2}{*}{Network} & \multicolumn{3}{c||}{MSE} & \multirow{2}{*}{Epochs} \\ \cline{2-4}
                         & \multicolumn{1}{c|}{u} & \multicolumn{1}{c|}{$\phi$} & \multicolumn{1}{c||}{$\theta$} & \\ \hline
$\mathcal N^{MOC}_{u, \phi, \theta}$ &\bf{0.7250} &0.0255 &0.0337 &794 \\ \hline
$\mathcal N^{MOC}_{\phi, \theta}$ &\_ &0.0212 &\bf{0.0289} &316 \\ \hline
$\mathcal N^{MOC}_{u}$ &0.7740 &\_ &\_ &721\\ \hline
$\mathcal N^{MOC}_{\phi}$ &\_ &\bf{0.0204} &\_ &248 \\ \hline
$\mathcal N^{MOC}_{\theta}$ &\_ &\_ &0.0305 &324 \\ \hline
\end{tabular}
\caption{MSE for neural networks trained on the MOC and QOC data sets. The networks differ on the output dimension.}
\label{tab:res}
\end{table}

\begin{figure}[t]
\sidecaption
\includegraphics[width=\linewidth]{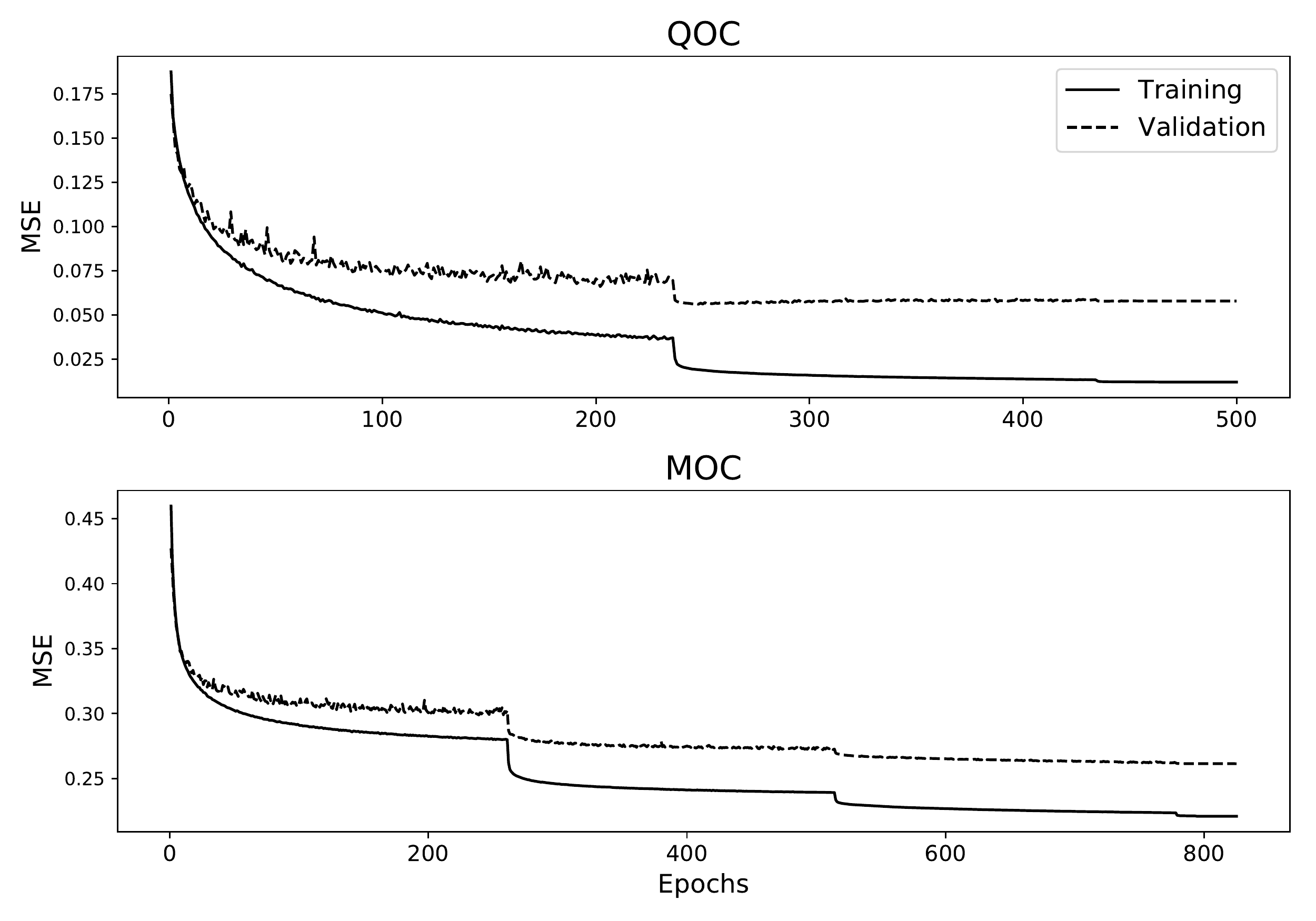}
\caption{Loss history for the networks $\mathcal N^{QOC}_{u, \phi, \theta}$ and $\mathcal N^{MOC}_{u, \phi, \theta}$. Note the jumps correspond to the learning rate being reduced.}
\label{fig:training_loss}
\end{figure}

\begin{figure}[t]
\sidecaption
\includegraphics[width=\linewidth]{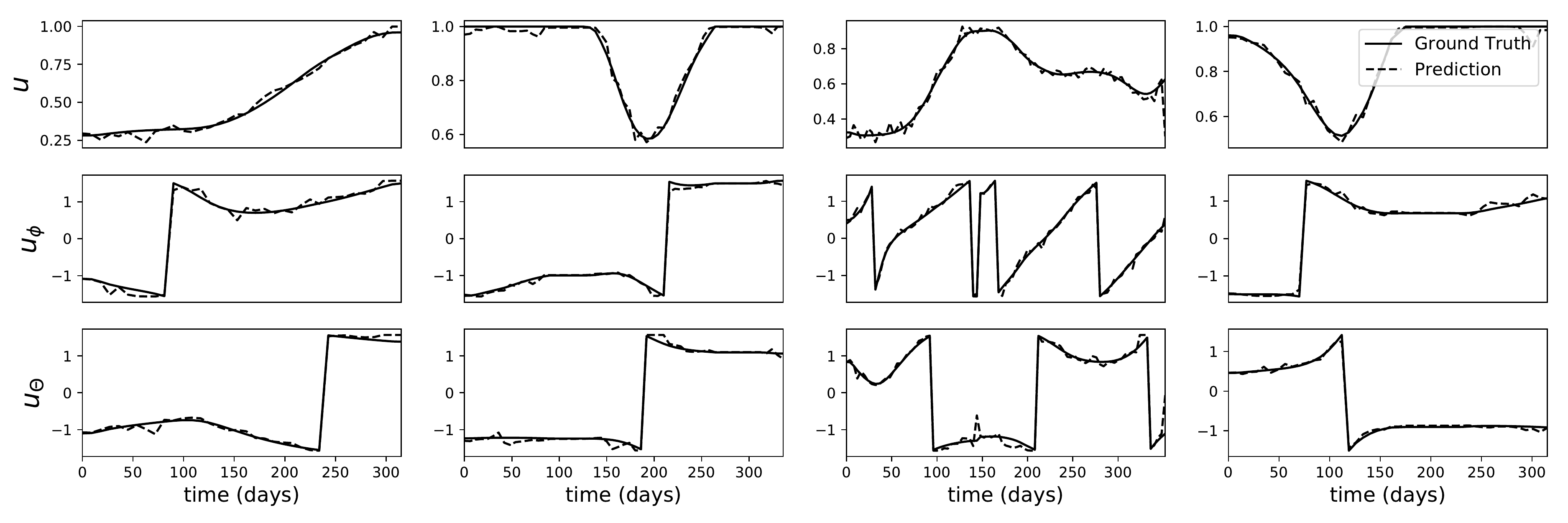}
\caption{Network $\mathcal N^{QOC}_{u, \phi, \theta}$ predictions on four different transfers from the QOC test set.}
\label{fig:prediction_qc}
\end{figure}

\begin{figure}[t]
\sidecaption
\includegraphics[width=\linewidth]{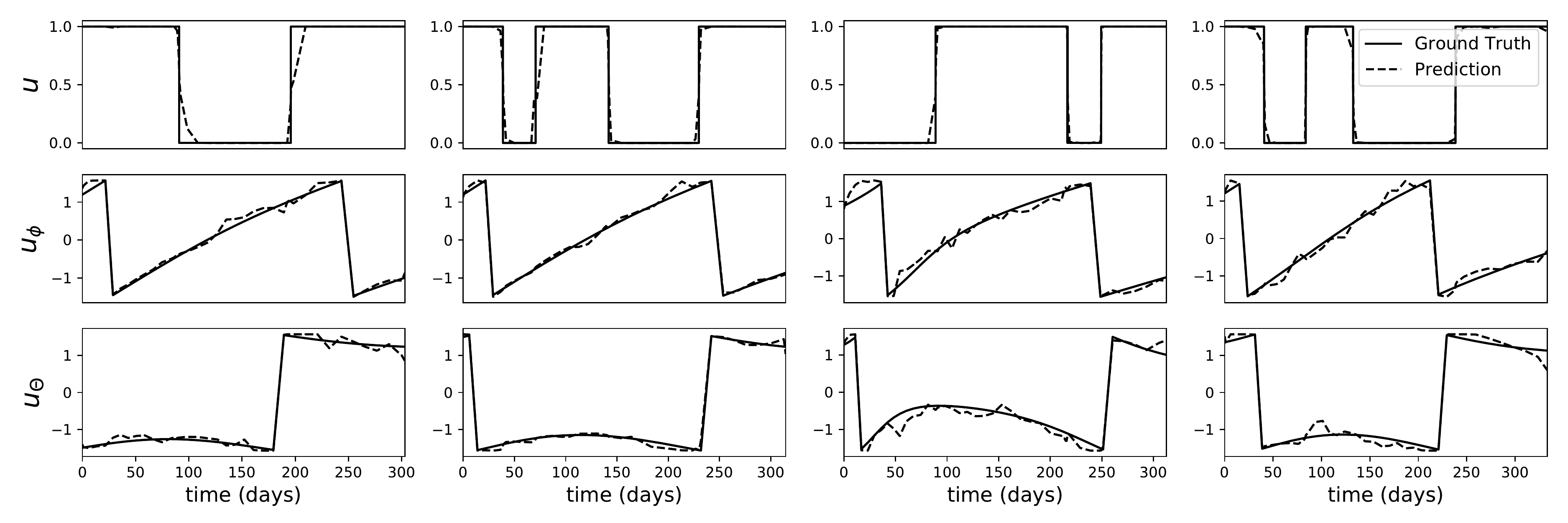}
\caption{Network $\mathcal N^{MOC}_{u, \phi, \theta}$ predictions on four different transfers from the MOC test set.}
\label{fig:prediction_moc}
\end{figure}

\section{Results}
We train different networks to predict the three outputs both concurrently or separately and on both $\mathcal{T}_{\mbox{MOC}}$ and $\mathcal{T}_{\mbox{QOC}}$ data sets. 
We indicate each network using superscripts and subscripts.
For example,  $\mathcal N^{MOC}_{\phi, \theta}$ indicates the network trained to model the variables $\phi$ and $\theta$ in the $\mathcal{T}_{\mbox{MOC}}$ data set. 
We report in Table \ref{tab:res} the obtained validation loss in the cases tested. We note that the best results for the output variables is obtained from different models indicating the absence of a clear trend to be exploited, while one could expect that predicting the entire thrust vector at once would bring advantages as the network would be able to share some of the weights as part of the necessary computations. We also report in Figure \ref{fig:training_loss} the loss during the training for the representative case of the $\mathcal N^{MOC}_{u, \phi, \theta}$ and $\mathcal N^{QOC}_{u, \phi, \theta}$ networks showing a relatively standard trend. We finally show four different optimal transfers and plot alongside them the optimal values against the predicted network values. The results are visualised in Figure \ref{fig:prediction_qc} and Figure  \ref{fig:prediction_moc}.

It appears that a deep network is able to represent the optimal guidance structure of the Earth-Mars transfer quite satisfactorily introducing errors that are, on average, rather small. 
One should also keep in mind that the actual quantitative value of the losses can be improved by further fine-tuning of the hyperparameters used in training the models.
In general, we observe the prediction of the throttle $u$ is better attained in the QOC case.
This is not surprising as the structure of the optimal control in the MOC case is highly discontinuous being a  bang-bang control.
We further note that data points close to the switching points are very difficult to predict correctly and when erroneously predicted result in big contributions to the overall loss.
These points form a majority in the datasets; a consequence of the time grid defined by the numerical integrator for the TPBVP shooting method solver.
This is the reason for the MSE on the $u$ prediction to be of a greater magnitude in the case of the MOC networks, while the visualised plot of the predictions for the same cases are actually returning a much better scenario.

A definitive assessment on the capability of the trained deep models to steer correctly the spacecraft has to be conducted, in a similar fashion to what done in \cite{sanchez2016real}, by integrating forward the whole spacecraft dynamics considering the deep network predictions as a control and will be part of a separate future publication.

\bibliographystyle{spbasic}
\bibliography{main}
\end{document}